\documentclass[conference]{IEEEtran}
\ifCLASSINFOpdf
\else
\fi
\usepackage[utf8]{inputenc}
\usepackage{cite} 
\usepackage{graphicx}
\usepackage{adjustbox}
\usepackage{amsmath}
\usepackage{amsfonts}
\usepackage{placeins}
\usepackage{fancyhdr}
\usepackage{balance}
\usepackage{algorithm,algpseudocode} 

\usepackage[dvipsnames]{xcolor}
 
\usepackage{amsthm}

\usepackage{lipsum}
\usepackage{subcaption}
\usepackage{mathtools}
\usepackage{booktabs}
\usepackage{threeparttable}
\usepackage{siunitx}

\begin{document}
%
\title{A Simulation Pipeline to Facilitate Real-World Robotic Reinforcement Learning Applications}

\author{
    \IEEEauthorblockN{Jefferson Silveira}
    \IEEEauthorblockA{\textit{Dept. of Electrical and Computer Engineering} \\
    \textit{Queen's University}\\
    Kingston, ON, Canada \\
    jefferson.silveira@queensu.ca}
    \and
    \IEEEauthorblockN{Joshua A. Marshall}
    \IEEEauthorblockA{
    \textit{Ingenuity Labs Research Institute} \\
    \textit{Queen's University}\\
    Kingston, ON, Canada \\
   joshua.marshall@queensu.ca}
    \and
    \IEEEauthorblockN{Sidney N. Givigi, Jr}
    \IEEEauthorblockA{\textit{School of Computing} \\
    \textit{Queen's University}\\
    Kingston, ON, Canada \\
    sidney.givigi@queensu.ca}
}


\maketitle

\begin{abstract}
Reinforcement learning (RL) has gained traction for its success in solving complex tasks for robotic applications. However, its deployment on physical robots remains challenging due to safety risks and the comparatively high costs of training. To avoid these problems, RL agents are often trained on simulators, which introduces a new problem related to the gap between simulation and reality. This paper presents an RL pipeline designed to help reduce the reality gap and facilitate developing and deploying RL policies for real-world robotic systems. The pipeline organizes the RL training process into an initial step for system identification and three training stages: core simulation training, high-fidelity simulation, and real-world deployment, each adding levels of realism to reduce the sim-to-real gap. Each training stage takes an input policy, improves it, and either passes the improved policy to the next stage or loops it back for further improvement. This iterative process continues until the policy achieves the desired performance. The pipeline's effectiveness is shown through a case study with the Boston Dynamics Spot mobile robot used in a surveillance application.  The case study presents the steps taken at each pipeline stage to obtain an RL agent to control the robot's position and orientation.
\end{abstract}
\IEEEpeerreviewmaketitle

\section{Introduction}
\label{sec:intro}

In recent years, Reinforcement Learning (RL) has achieved high-level performance in solving challenging robotic tasks. 
While RL itself is not new (with roots dating back to the 1960s~\cite{minsky1961steps}), its recent successes can be attributed to several factors: the availability of powerful and affordable processing units (such as CPUs and GPUs), advancements in deep learning techniques, and the development of new deep RL methods.

The RL methodology relies on three core concepts: the agent, the reward signal and the environment. The agent is the decision-maker that learns by interacting with the environment. The reward signal defines the goal of the RL problem, serving as feedback to encourage desirable behaviours and discourage undesirable ones. The environment is where the agent operates, providing observations in response to the agent's actions~\cite{sutton2018reinforcement}. The agent gathers experience to learn the appropriate actions and solve the RL problem. This is an iterative approach that can be directly applied to a physical robot. 



Unfortunately, applying RL directly in real-life applications is often prohibitive, requiring the robot to collect data by iteratively acting on the environment~\cite{levine2020offline}. This iterative process can be costly and time-consuming. There may be safety concerns because the robot may take unexpected actions during training, possibly causing damage to itself or its surroundings.

\begin{figure*}
    \centering
    \includegraphics[width=0.85\linewidth]{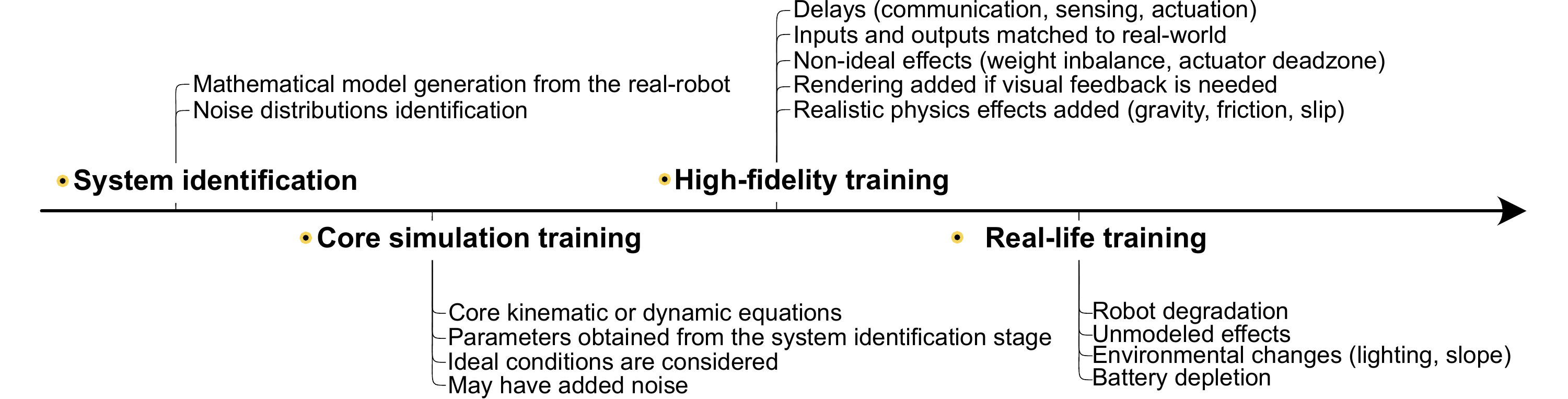}
    \caption{Simplified diagram describing the components of the proposed RL pipeline. Each component is optional, and the combination of the stages depends on the problem's complexity. 
    }
    \label{fig:simplified_diagram}
\end{figure*}

The success of RL in robotics is partly due to the development of a methodology known as off-policy RL, which enables offline training~\cite{levine2020offline} where the RL training process happens after data collection. Algorithms such as Q-learning~\cite{sutton2018reinforcement}, SAC~\cite{haarnoja2018soft}, and TD3~\cite{fujimoto2018addressing} are examples of off-policy RL. Another advantage of off-policy RL is that it allows for the use of simulated data during training. Unfortunately, this approach introduces other challenges, because discrepancies between the simulated and the real world may cause the robot to perform worse when compared to its simulated counterpart. This problem, often referred to by various terms such as the reality gap, sim-to-real gap, and sim2real gap~\cite{jakobi1995noise}, arises because off-policy RL is usually trained mostly, if not entirely, in simulation. Some studies focus on reducing this problem~\cite{chebotar2019closing, calderon2024deep, zhao2020sim}, and strategies they apply include (\textit{i}) creating high-fidelity simulations with parameters modelled directly from the robot, and (\textit{ii}) randomizing the simulation parameters to generate a policy that is robust to environmental variations. 

Considering these methodologies, this work presents a pipeline to facilitate the development of RL models on real-world robotic systems. 
A simplified overview of each phase's components is shown in Fig.\ \ref{fig:simplified_diagram}. 
The proposed pipeline combines established methodologies from the RL literature into a staged process that involves system identification, training with different levels of simulation complexity, and real-world deployment. 
~While the individual techniques included in the pipeline are not novel, this paper's contribution lies in a systematic approach to the learning process. Unlike most studies, which often detail only the specific steps for their application, this pipeline offers a modular framework to guide practitioners in adapting RL models to real-world scenarios. To support this, a case study demonstrating the successful deployment of an RL policy on a mobile robot is also presented.

The remainder of the paper is organized as follows. Section~\ref{sec:related} presents common approaches to reduce the reality gap; Section \ref{sec:pipeline} describes the proposed pipeline and discusses its application in robotics; Section \ref{sec:case_study} describes the steps taken to train an RL policy using the proposed pipeline; Section \ref{sec:application} illustrates the results in a surveillance application. Finally, Section \ref{sec:conclusion} presents the final remarks.

\section{Related Work}
\label{sec:related}

This section presents previous studies on the simulation-to-reality gap and links these concepts to the proposed pipeline. For this purpose, it is necessary to define the concept of environment used in this paper. In robotic RL applications, the environment includes not only the space where the robot operates but also the robot itself, and the agent is the decision-making algorithm or policy that sends commands via the robot.  However, it is also common to refer to the robot along with the learning algorithm as the agent and the environment as everything external to the robot. This paper adopts this latter convention to distinguish the robot from its surroundings.

Research on RL in robotic systems has led to various methodologies for improving sample efficiency of RL algorithms and reducing the sim-to-real problem. The sample efficiency in RL refers to the number of samples (i.e., experiences) needed to achieve a certain performance level. A comprehensive survey on the topic is presented by Calderón-Cordova et al.~\cite{calderon2024deep}. They present many techniques, frameworks, tools, and a practical guide for developing RL control applications focused on robotic manipulators. Even though their work is extensive, they propose a simple pipeline containing a single simulation stage with no discussion of the various levels of simulation complexity and their effects on the reality gap. 

In contrast,~\cite{zhu2021survey} presents a survey on RL applied to bio-inspired robots that categorize RL methodologies into four main groups: 1) methods that rely on accurate simulators; 2) approaches that only use simplified kinematic or dynamic models; 3) techniques that apply RL on top of hierarchical controllers; and 4) methods that leverage human demonstrations. They also mention that while methodologies in group 1 often perform better after a sim-to-real transfer, they are less sample-efficient than the others. Our proposed pipeline supports not only a single simulation stage, as presented in~\cite{calderon2024deep} but also allows for multi-level simulation complexities. This includes options to integrate high-fidelity simulators and simplified kinematic or dynamic models as core simulators.

In another survey, Zhao et al.~\cite{zhao2020sim} discuss not only the potential benefits of high-fidelity simulation in improving the sim-to-real transfer but also present three other approaches: system identification to create a simulator tailored to a specific robot, domain randomization, and domain adaptation. Domain randomization methods involve modelling parameters from reality and randomizing their values in the simulator to cover the actual distribution of these parameters in the real world. On the other hand, domain adaptation involves the combination of two or more environments during the training process. For example, one could train a model in a high-fidelity simulator and then transfer the model to training on real-world data.

Several works successfully use domain randomization~\cite{huber2024domain,peng2018sim,akkaya2019solving}. In~\cite{huber2024domain}, a feedback approach is used to change the simulation parameters based on how the real robot performs with the transferred policy, thus allowing for automatic randomization to achieve high performance on the transferred policy. In contrast,~\cite{peng2018sim} performs a comprehensive randomization of parameters (e.g., table size, arm dynamics, controller gains, links' mass, friction, noise levels, and time steps) on a robotic arm application, 
totalling 95 randomized parameters, resulting in a successful sim-to-real transfer without additional training on the physical system. Using a combination of system identification techniques, domain randomization and curriculum learning, \cite{akkaya2019solving} obtained a remarkable level of dexterity in a robotic hand when trained to solve the Rubik's cube.


Domain adaptation techniques involve converting input data from one domain into another, where most training is performed. For example,~\cite{james2019sim} develops a machine-learning model that converts the visual feedback data from the real world into a format that resembles simulated inputs. This translation technique allows a model trained in simulation to perform well on physical robots. Similar strategies are used in~\cite{bousmalis2018using}.

Note that these methodologies are not always necessary. For example,~\cite{hu2021sim} presents a sim-to-real pipeline to address the challenge of robot navigation in 3D cluttered environments. Their pipeline includes a simulation stage that matches the inputs and outputs to the real robot, and uses simulated sensors and state estimation techniques that closely matched the real ones. These steps achieved successful real-world transfer without requiring any adaptation or additional training. While control problems like these could be solved with classical control techniques, RL provides a data-based adaptive methodology that is independent of the robot model.

Ultimately, the requirements of the RL process depend on the complexity of the task. If the robot is passively stable and accepts simple commands, such as linear and angular velocities, it is possible to transfer the learned model without further adaptation~\cite{hu2021sim}. However, more complex problems, such as solving a Rubik's cube with 24 degrees of freedom, require more steps to achieve acceptable performance in real applications~\cite{akkaya2019solving}. 

\section{Proposed Reinforcement Learning Pipeline}
\label{sec:pipeline}

\begin{figure*}
    \centering
    \includegraphics[width=1\linewidth, trim=0cm 0cm 0cm 0.4cm, clip]{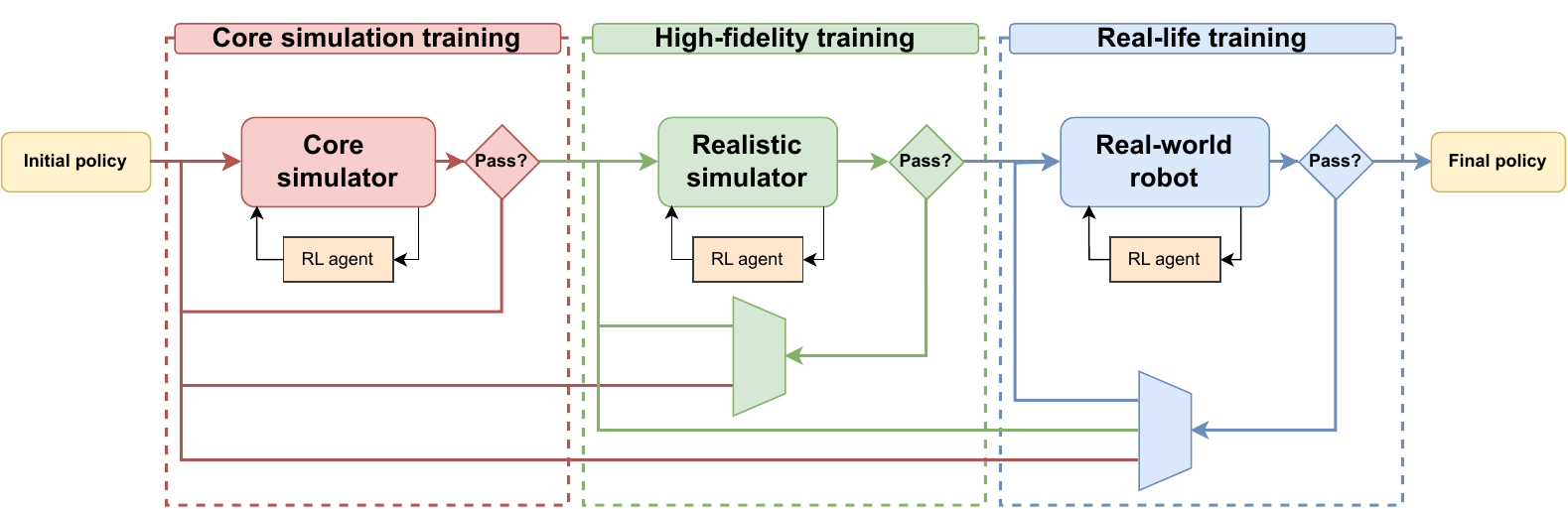}
    \caption{The proposed training pipeline involves three stages with increasing levels of complexity from left to right. Each stage is optional and can be revisited with different parameters until they pass a predefined performance criteria. 
    Each stage receives as input a policy and outputs a modified policy, allowing incremental improvement until the final policy is achieved.}
    \label{fig:training_pipeline}
\end{figure*}

The pipeline is proposed as a multi-stage process to turn RL policies into real-world applications as presented in Fig.~\ref{fig:simplified_diagram}, with four phases: system identification, core simulation training, high-fidelity training, and real-life training, which are explained further in this section.
Fig.\ \ref{fig:training_pipeline} describes the data flow and interactions between the RL agent and the environments during training. This design allows the customization of each phase to meet the specific requirements of a task. For example, tasks that involve manipulating physical objects (e.g., a Rubik’s cube~\cite{akkaya2019solving}) may benefit from utilizing all stages of training. In contrast, more straightforward tasks with no interaction with other objects, like the position control of a wheeled vehicle, might only require the system identification and core simulation stages. This flexibility is valuable, especially for researchers and industry practitioners new to RL.

The following subsections provide a detailed description of each component and their interactions in the proposed pipeline.



\subsection{System Identification}

This stage focuses on obtaining a data-based model of the robot to be applied in the following stages. Although optional, this step can be crucial to reduce the reality gap by considering the robot's physical parameters in the simulators used in the following stages. The system identification field is vast, and techniques include identifying linear or nonlinear models that convert inputs to outputs, frequency response analysis, machine learning, and many others. 
A good resource on system identification is provided in~\cite{brunton2022data}. Section \ref{sec:case_study} provides further details on the system identification technique used in the presented case study. 

\subsection{Core Simulation Training}

This is the first training stage in Fig.\ \ref{fig:training_pipeline}, and it allows the RL agent to learn within a simplified simulation environment. This phase could involve modelling either the kinematic or dynamic equations of motions. For example, an RL agent could be trained to control a differential-drive robot by modelling the kinematic equations while ignoring factors like friction, motor mismatch, and other sources of error, such as in \cite{hu2021sim}, where RL is applied to control a wheeled robot to navigate in rough terrain. While this simplified approach might allow the policy to transfer to the actual robot with minimal training, the transfer could also result in policy degradation. Incorporating system identification parameters into the simulation could ensure smoother transfer and more reliable performance.

The passing criteria in this and subsequent stages can be a numerical function that determines if the model passes a predefined performance score or based on practitioners' prior experience to assess model feasibility. Examples of performance metrics include the success rate for goal-based problems or the accumulated reward.
The tools used in this stage include the gymnasium API developed for creating RL simulators and physics simulators, such as Bullet and MuJoCo, that can be used to implement rigid body dynamics.

\subsection{High-Fidelity Training}

The second training stage in Fig.\ \ref{fig:training_pipeline} involves high-fidelity simulation, incorporating realism such as gravity, friction, actuator latency and dead zones, sensor noise, and realistic renderings. This reduces the sim-to-real gap, critical for complex RL tasks. With high-fidelity simulators, domain randomization is a powerful technique to further reduce the sim-to-real gap. For example, domain randomization (e.g., cube size, friction, force ranges, inertia, and action latency) enhanced the training process in the Rubik's cube robot manipulation task~\cite{akkaya2019solving}, where it enabled successful transfer into the real robot.

Common tools used in this stage, such as Gazebo, CoppeliaSim and Isaac Sim, offer advanced features like accurate sensor modelling, real-robot input/output matching and Robot Operating System (ROS) support, enabling near-seamless transitions between simulated and real environments.

\subsection{Real-Life Training}

In the final phase, the RL model is deployed on the physical robot, and the performance degradation is evaluated. If performance is inadequate, fine-tuning the model with real-world data or addressing discrepancies through high-fidelity simulation, using either domain randomization or domain adaptation, can reduce the sim-to-real gap. This iterative process continues until the model meets the desired performance criteria.


\subsection{Debugging RL Applications with the Pipeline}

RL requires careful integration of components, including actions, observations, reward signals, simulation, RL algorithms, and well-tuned hyperparameters. Complex robotics tasks with continuous observations and actions often require multiple neural networks, further increasing the number of hyperparameters that must be optimized.

More often than not, the initial training fails to converge to a useful policy, with issues stemming from suboptimal hyperparameters, insufficient observations, or neural networks that need more neurons or layers. In these situations, a common solution is to start with a simplified simulation and minimal observations and actions, making it easier to find workable parameters. This process can then iterate, gradually increasing task complexity or progressing through the pipeline stages until the desired performance is obtained.

\section{Case Study on a Mobile Robot}
\label{sec:case_study}

This section presents a case study on applying the discussed concepts to obtain an RL policy for the Boston Dynamics Spot robot. Spot, an agile legged robot, autonomously computes gait and foot placement on linear ($a_x$), lateral ($a_y$) and angular body velocities ($a_\theta$) commands. The objective was to train an RL model to control the robot's position and orientation to reach a desired configuration while optimizing a cost function. This controller was then used in a surveillance application.

\subsection{System Identification}
\label{sec:system_id}

In this stage, the objective was to reduce the sim-to-real gap by modelling the robot's motion constraints. The robot cannot perfectly execute commanded velocities because converting desired speeds into leg movements and actuation limits introduces errors. Training an RL agent solely on commanded velocities $\mathbf{a} = \left(a_x,a_y,a_\theta\right)$ risks poor control because these may not align with what the robot can physically achieve.

\begin{figure}
  \begin{subfigure}[b]{0.45\linewidth}
    \centering
    \includegraphics[width=1.0\linewidth, trim=5.2cm 1.3cm 3.2cm 2.0cm, clip]{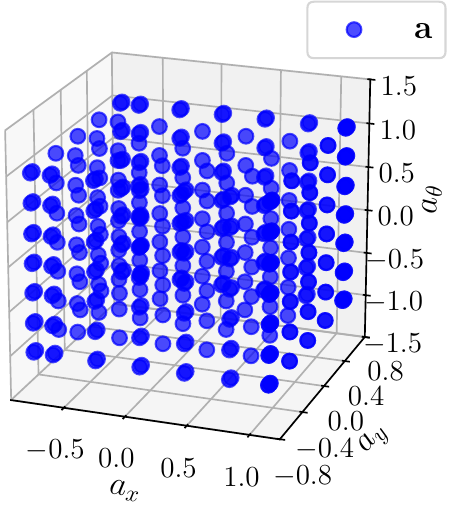}
    \caption{} 
    \label{fig:action_set_a} 
  \end{subfigure}
  \begin{subfigure}[b]{0.45\linewidth}
    \centering
    \includegraphics[width=1.0\linewidth, trim=5.0cm 1.3cm 3.4cm 1.0cm, clip]{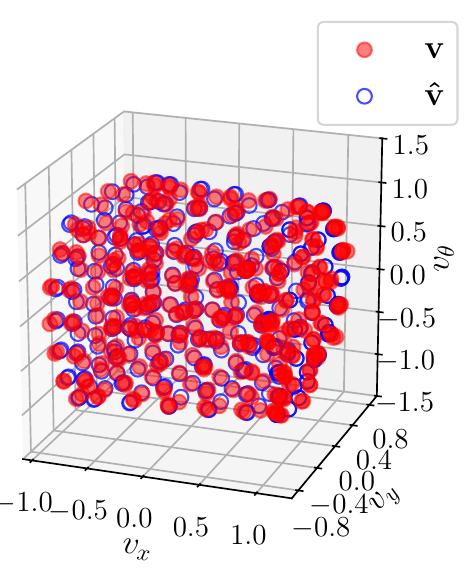} 
    \caption{} 
    \label{fig:action_set_b}
  \end{subfigure}
  \caption{Action set of the Spot robot. (a) Nominal and (b) feasible velocities with overlaid approximated velocities. }
  \label{fig:action_set} 
\end{figure}

The first step was identifying the velocities the robot could execute. To do so, a grid of commanded body velocities $\mathbf{a}$  was created (Fig.\ \ref{fig:action_set_a}), and the executed body velocities $\mathbf{v} = \left(v_x,v_y,v_\theta\right)$ were measured using a motion capture system, with grid ranges shown in Table \ref{tab:system_params}. Note that $a_x$ is not symmetrical because the robot moves faster forward than backward, adding control complexity, which the RL agent can still learn to handle.

\begin{figure}
\centering\includegraphics[width=0.9\linewidth]{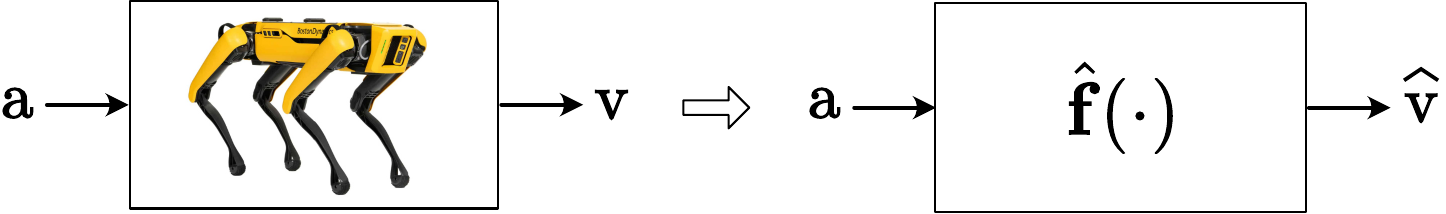}
    \caption{The executed velocities of the real robot are approximated using a polynomial function approximator.}
    \label{fig:fuction_approximation}
\end{figure}

The system identification process involved finding a function that approximates the executed velocities from commanded velocities (Fig.~\ref{fig:fuction_approximation}), represented as
\begin{equation}
    \hat{\mathbf{v}} = \hat{\mathbf{f}}(\mathbf{a}),
\end{equation}
where  $\hat{\mathbf{v}}$ is the approximated body velocity and $\hat{\mathbf{f}}(\mathbf{a})$ is a multivariate function with a third-order polynomial with no bias term in each dimension of $\mathbf{a}$, resulting in:

\begin{equation}
    \hat{\mathbf{v}} = \begin{bmatrix}
           \hat{v}_x \\
           \hat{v}_y \\
           \hat{v}_{\theta}
         \end{bmatrix}
     =
     \begin{bmatrix}
         \hat{f}_x(\mathbf{a})\\
         \hat{f}_y(\mathbf{a})\\
         \hat{f}_\theta(\mathbf{a})
     \end{bmatrix},
\end{equation}
where $\hat{f}_x(\mathbf{a})$, $\hat{f}_y(\mathbf{a})$, and $  \hat{f}_\theta(\mathbf{a})$ are the polynomial functions that approximate the executed velocities in each dimension. For clarity, the approximation for $\hat{v}_x$ is given by
\begin{equation}
    \hat{v}_x = \hat{f}_x(\mathbf{a}) = \sum_{0 < i+j+l \leq 3} c_{x,ijl} (a^i_x a^j_y a^l_\theta),
\end{equation}
 where $i, j, l \in \mathbb{Z}_{\geq 0}$ and the coefficients $c_{x,ijl}$ are computed using the least squares method that minimizes the total squared error
\begin{equation}
    J_x = \sum_{n=0}^{N-1} (v_{x,n} - \hat{v}_{x,n})^2,
\end{equation}
where $N$ represents the total number of samples in the grid (Fig. \ref{fig:action_set_a}), and the $n$ index represents the $n$-th element. The coefficients of $\hat{v}_y$ and $\hat{v}_{\theta}$ are computed similarly. 

Fig.\ \ref{fig:action_set_b} shows the linear regression results with the approximated velocities overlaid on the executed ones, illustrating a good fit from the polynomial regression.

\subsection{RL Formulation}

The problem was modelled as a goal-conditioned Markov Decision Process (MDP) $\langle S, G, A, T, R \rangle$, where $S$ is the state space, $G$ the goal space, $A$ the action space, $T: S\times A \rightarrow S$ the state transition function, and $R(\mathbf{a}, \mathbf{s}, \mathbf{g}),~R: S \times A\rightarrow\mathbb{R}$ the reward function that returns a scalar when taking action $\mathbf{a}$ and arriving at state $\mathbf{s}$ with goal $\mathbf{g}$. The goal is to find a policy $\pi(\mathbf{a}|\mathbf{s},\mathbf{g})$ that maximizes the agent's cumulative reward~\cite{antonyshyn2023multiple}. The policy $\pi(\mathbf{a}|\mathbf{s},\mathbf{g})$ defines a probability distribution over the action space $A$ given a combination of state $\mathbf{s}$ and goal $\mathbf{g}$. The policy can be applied probabilistically, where each action $\mathbf{a}$ is a sample of the probability distribution, or deterministically, where $\mathbf{a}$ is the distribution mean.

\subsubsection{State and Goal Spaces}

Since the goal of the RL agent is to control the position and orientation of the robot, the state was defined as
\begin{equation}
    \mathbf{s} = (x, y, \theta)\in\mathbb{R}^2\times[-\pi,\pi).
\end{equation}
Similarly, the goal is defined as
\begin{equation}
    \mathbf{g} = (x_g,y_g,\theta_g) \in[r_{\rm min},r_{\rm max}]^2\times[-\pi,\pi),
\end{equation}
where $r_{\rm min}$ and $r_{\rm max}$ represent the lower and upper limits of $x_g$ and $y_g$. Note that the state and goal combination form the observation $\mathbf{o} = (\mathbf{s},\mathbf{g})$, but separating them into state and goal vectors is more intuitive.

\subsubsection{Action Space}

The commanded action was defined as 
\begin{equation}
    \mathbf{a} = (a_x, a_y, a_\theta) \in \mathbb{R}^3.
\end{equation}
They represent the linear, lateral and angular body velocities commands sent to the robot. 

\subsubsection{Simulation and Transition Function}

The simulator used in the core simulation stage converts the desired action at time step $k$ into the identified executed action, as shown in Section \ref{sec:system_id}, and computes a new state resulting from the applied action. The new state is obtained through a first-order integration of the executed velocity as
\begin{align}
    \hat{\mathbf{v}}_k &= \hat{f}(\mathbf{a}_k) \\
    \mathbf{s}_k &= \mathbf{s}_{k-1} + \hat{\mathbf{v}}_k\Delta t,
\end{align}
with $\Delta t$ being the step duration.

Since this is a goal-conditioned MDP, our simulator must also inform when the robot reaches $\mathbf{g}$. To do so, the error in position and orientation are computed
\begin{align}
    e_{p,k} &= \|\mathbf{p}_g-\mathbf{p}_{s,k}\|_2, \\
    e_{\theta,k} &= |\theta_g - \theta_{k}|,
\end{align}
where $\mathbf{p}_g = (x_g, y_g)$, $\mathbf{p}_{s,k} =(x_k, y_k)$, and $||\cdot||_2$ represents the Euclidean norm. Then, the robot successfully reaches $\mathbf{g}$ when $e_{p,k} < \epsilon_p$ and $e_{\theta,k} < \epsilon_\theta$, where $\epsilon_p$ and $\epsilon_\theta$ are tolerance parameters that define the required proximity to the goal.

\subsubsection{Reward Function}

The reward function for this study was chosen based on common costs used in model predictive control applications. For interested readers on this topic, the authors of \cite{song2023reaching} present an insightful comparison between optimal control and reinforcement learning techniques and their costs. Our objective was to optimize a policy that minimizes control actions, action smoothness, and time, resulting in the following cost function:
\begin{equation}
\label{eq:cost_function}
J = \sum_{k=0}^{N-1}\left(\|\mathbf{u}_k\|_{\mathbf{R}}^2 + \|\mathbf{u}_k-\mathbf{u}_{k-1}\|_{\mathbf{S}}^2 + \lambda_k\right),
\end{equation}
where the notation $\|\mathbf{z}\|^2_\mathbf{M}= \mathbf{z}^\top \mathbf{M} \mathbf{z}$ represents the weighted Euclidean norm, with $\mathbf{z}$ as a column vector and $\mathbf{M}$ a square matrix matching the dimension of $\mathbf{z}$. The parameters $\mathbf{R}$, $\mathbf{S}$ prioritize action magnitude and variation, respectively, while $\lambda_k$ is a time-varying factor that encourages faster task completion. For example, the parameters $\lambda_k = 1$ and $\mathbf{R} = \mathbf{S} = \mathbf{0}$ define the minimum time optimal control problem \cite{rosolia2021minimum}.

Since RL maximizes a reward rather than minimizing a cost, the cost function in \eqref{eq:cost_function} is converted into a reward as follows,
\begin{equation}
\label{eq:reward}
    r_k =  -\left(\|\mathbf{u}_k\|_{\mathbf{R}}^2 + \|\mathbf{u}_k-\mathbf{u}_{k-1}\|_{\mathbf{S}}^2 + \lambda_k\right).
\end{equation}

Maximizing \eqref{eq:reward} over time corresponds to minimizing the cost in \eqref{eq:cost_function}. However, to further encourage the RL agent to reach the goal state, the time penalty is removed once the robot is within a specified proximity to the goal. Thus, we define:
\begin{equation}
\lambda_k = 
\begin{cases} 
      0, & \text{if }  e_{p,k} < \epsilon_p \text{ and } e_{\theta,k} < \epsilon_\theta \\ 
      1, & \text{otherwise} 
   \end{cases}
\end{equation}

\subsubsection{Neural Network Architecture}

\begin{figure} 
    \centering
    \includegraphics[width=1.0\linewidth, trim=0cm 0cm 0cm 0cm, clip]{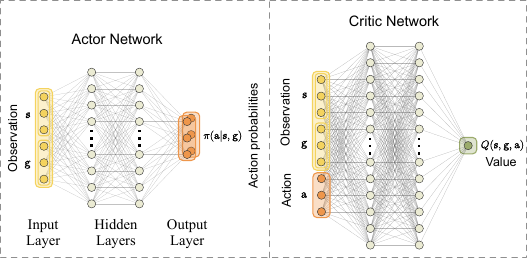}
  \caption{Actor-critic network architecture}
  \label{fig:actor_critic_architecture} 
\end{figure}

In recent years, several algorithms, like DDPG~\cite{lillicrap2015continuous}, SAC~\cite{haarnoja2018soft}, and TD3~\cite{fujimoto2018addressing}, have been created for continuous state and action spaces.  Considering the pool of deep RL algorithms, we chose the Soft Actor-Critic Algorithm (SAC)~\cite{haarnoja2018soft} due to being easy to find hyperparameters that lead to good policies. The stochastic nature of SAC also helps with environment exploration, which reduces the number of parameters to tune. The RL training parameters and network architectures used are listed in Table~\ref{tab:system_params}, and a visual representation of the network architecture for the actor-critic framework is presented in Fig.\ \ref{fig:actor_critic_architecture}.

\begin{figure*}
  \begin{subfigure}[b]{0.33\linewidth}
    \centering
    \includegraphics[width=1.0\linewidth, trim=0.3cm 0.3cm 0.2cm 0.1cm, clip]{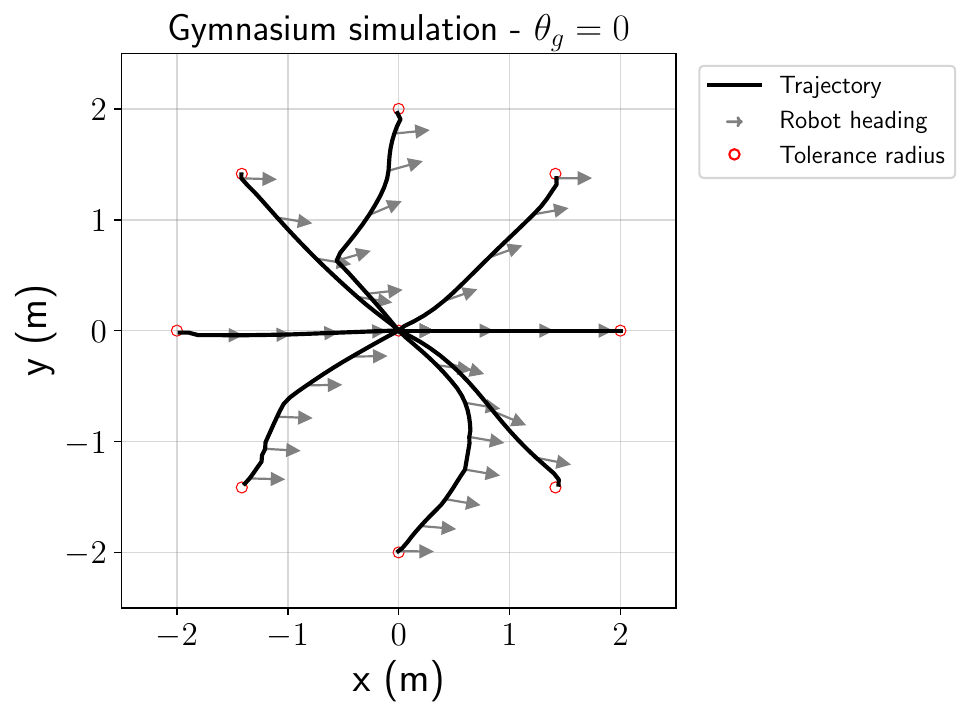}
    \caption{Core simulation} 
    \label{fig:policy_results_a} 
  \end{subfigure}
  \begin{subfigure}[b]{0.33\linewidth}
    \centering
    \includegraphics[width=0.71\linewidth, trim=0.3cm 0.3cm 0.2cm 0.1cm, clip]{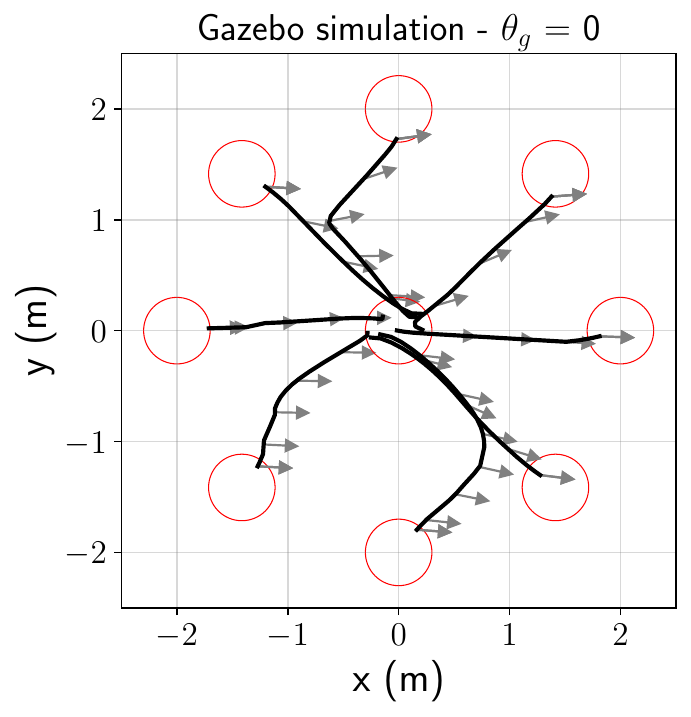} 
    \caption{High-fidelity simulation} 
    \label{fig:policy_results_b}
  \end{subfigure}
 \begin{subfigure}[b]{0.33\linewidth}
    \centering
    \includegraphics[width=0.71\linewidth, trim=0.3cm 0.3cm 0.2cm 0.1cm, clip]{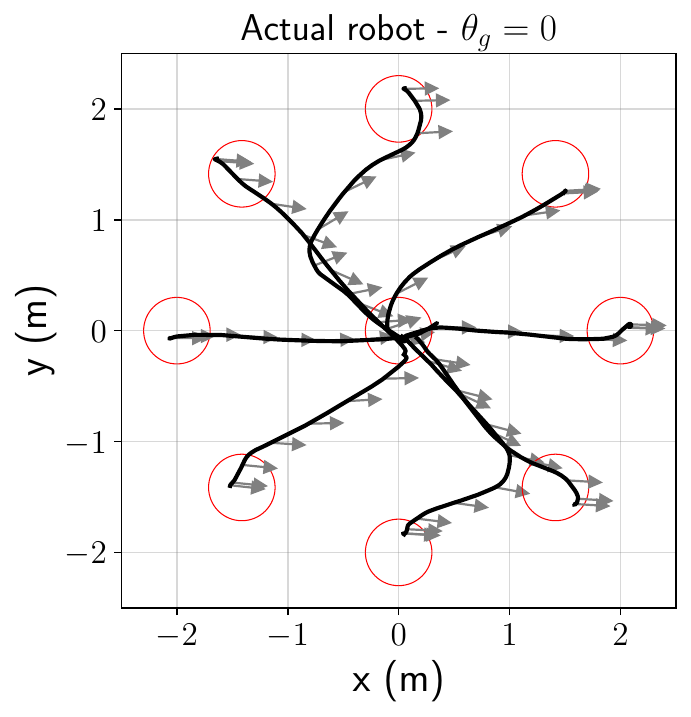}
    \caption{Robot execution} 
    \label{fig:policy_results_c} 
  \end{subfigure}
  \caption{Execution of the policy at multiple goals with $\theta_g = 0$ with (a) showing the core simulation trajectories, (b) the high-fidelity simulation trajectories, and (c) the trajectories executed by the real robot.}
  \label{fig:policy_results} 
\end{figure*}
In SAC, two networks are used: the actor and the critic. The actor network is the policy itself, taking the observation (state and the goal) as input and outputting action probabilities in terms of mean and deviation for each element. The critic network takes both the observation and a sampled action as input, and outputs estimates of the expected return of that action given the current policy (Q-value). This Q-value guides the actor’s updates, helping SAC refine the policy by assessing the value of actions to improve the long-term rewards.

\begin{table}
\caption{Simulation, RL, and neural network parameters.}
\begin{center}
\begin{threeparttable}
\begin{tabular}{l c}
    \toprule
     \textbf{Parameter} & \textbf{Value}\\ \midrule
    
    Simulation frequency & $30$ Hz \\
    Policy frequency & $10$ Hz \\
    Observation standard deviation & $0.01$ \\
    $\mathbf{R}$ & $\textup{diag}(0.0, 0.8, 0.8)$\\
    $\mathbf{S}$ & $\textup{diag}(0.2, 0.2, 0.2)$\\
    Range of $a_x$ & $[-0.8, 1.1]$ m/s\\
    Range of $a_y$ & $[-0.7, 0.7]$ m/s\\
    Range of $a_\theta$ & $[-1.1, 1.1]$ rad/s\\
    $[r_{\rm min}, r_{\rm max}]$ & $[-2,2]$ \\
    \midrule
   
    Batch size   &   $512$\\
    $\tau$ & $0.0045$ \\
    Discount factor   &  $0.999$\\
    Learning rate   &   \SI{2e-4}{} \\
    Buffer size   &  \SI{1e6}{} \\
    Number of training steps & \SI{3e5}{} \\
    \midrule
    Actor neurons & $16$ \\
    Actor hidden layers & $2$\\
    Critic neurons & $128$ \\
    Critic hidden layers & $2$ \\
    Hidden layer activation function & ReLu\\
    Output layer function & linear \\

    \bottomrule
\end{tabular}
\end{threeparttable}
\end{center}
\label{tab:system_params}
\end{table}  

\subsection{Training and Verification Process}

The core simulation training stage was performed by using Gymnasium~\cite{towers2024gymnasium}, which is a library that simplifies the creation of simulators tailored for RL applications, and the SAC neural network was sourced from the stable-baselines 3 library~\cite{stable-baselines3}. The simulation parameters are shown in Table~\ref{tab:system_params}. To accelerate learning, another concept called Hindsight Experience Replay (HER)~\cite{andrychowicz2017hindsight} was also applied. HER is a technique that generates additional training data from failed episodes by redefining the goal to a point the robot reached and adjusting rewards accordingly. This approach enables the agent to learn not only from successes but also from failures.

Two additional training strategies were applied in this study: curriculum learning and randomization. Curriculum learning was applied to the goal tolerances $\epsilon_p$ and $\epsilon_\theta$, starting with the tolerances covering $80$~\% of the training region. When the robot reached a $95$~\% success rate, the tolerances were reduced by $20$~\% until it reached $\epsilon_p = 0.05$ m and $\epsilon_\theta = \ang{1}$. Looking at the pipeline in Fig.\ \ref{fig:training_pipeline}, this can be viewed as following the feedback loop in the core simulator until the RL agent achieves the desired tolerances. Randomization was applied to the robot's state, introducing noise in the robot's position and orientation to encourage the RL agent to learn how to control the robot under uncertain conditions, reducing the reality gap.

Once the policy achieved a $100$~\% success rate in the core simulator stage with the final tolerance levels, it was transferred to the high-fidelity simulation stage using the Gazebo simulator, which provides a physics engine, realistic sensor emulation, and a communication layer with ROS. Because the problem did not involve interacting with physical objects, there was no necessity to continue training the policy on the Gazebo simulator. However, running the policy on Gazebo was still crucial, as integration with ROS enabled the use of the same code, and localization and navigation architecture as on the real robot, allowing the algorithm to be tested under conditions that closely replicate the real-world environment. 

After testing the RL agent in Gazebo, the policy was deemed ready for deployment on the real robot. Initial deployments yielded positive results, though some discrepancies were observed compared to the Gazebo simulation. Specifically, the RL agent struggled to stop the robot in certain scenarios, such as the goal at $x = 0$ m and $y = 2$ m in Fig.~\ref{fig:policy_results_c}., leading to overshooting and oscillations around the goal position. This was caused by the robot's latency in stopping due to inertia and a minimum action duration not accounted for in the core simulator. Thus, the tolerance to reach the goal was increased to $\epsilon_p = 0.3$ m and $\epsilon_\theta = \ang{17}$. This modification eliminated the oscillations and proved sufficient for our application, allowing us to avoid further training. The need for increased tolerances is a consequence of the reality gap. If more precise positioning is required, the pipeline allows for more training using either real-life data during training or by incorporating the robot oscillations in the high-fidelity simulator.

Fig.~\ref{fig:policy_results} shows the robot trajectories across all three stages. The  Gymnasium simulation (Fig.\ \ref{fig:policy_results_a}) shows accurate trajectories, as anticipated. Note that trajectories requiring lateral movement include some  forward/backward motion due to the RL agent being optimized for faster executions rather than shorter trajectories. In the Gazebo runs (Fig.\ \ref{fig:policy_results_b}), the tolerances were adjusted to match those used in the actual robot stage, and minor variations in the robot's starting positions were introduced to create a more realistic setup. Despite these differences, the trajectories were similar to the Gymnasium stage. Finally, in the real robot execution (Fig.\ \ref{fig:policy_results_c}), the trajectories closely matched the Gazebo results, except for a delay in the robot stopping upon reaching the goal regions.

\section{Application in Surveillance}
\label{sec:application}

The trained RL agent was tested on the Boston Dynamics Spot robot in a surveillance task conducted inside one of Queen's University buildings. The robot was tasked with safely visiting multiple waypoints within the building using the navigation system shown in Fig.~\ref{fig:navigation_system}.

\subsection{Description of the Navigation System}
\begin{figure}
    \centering
    \includegraphics[width=1.0\linewidth, trim=0.0cm 0cm 0.0cm 0.0cm, clip]{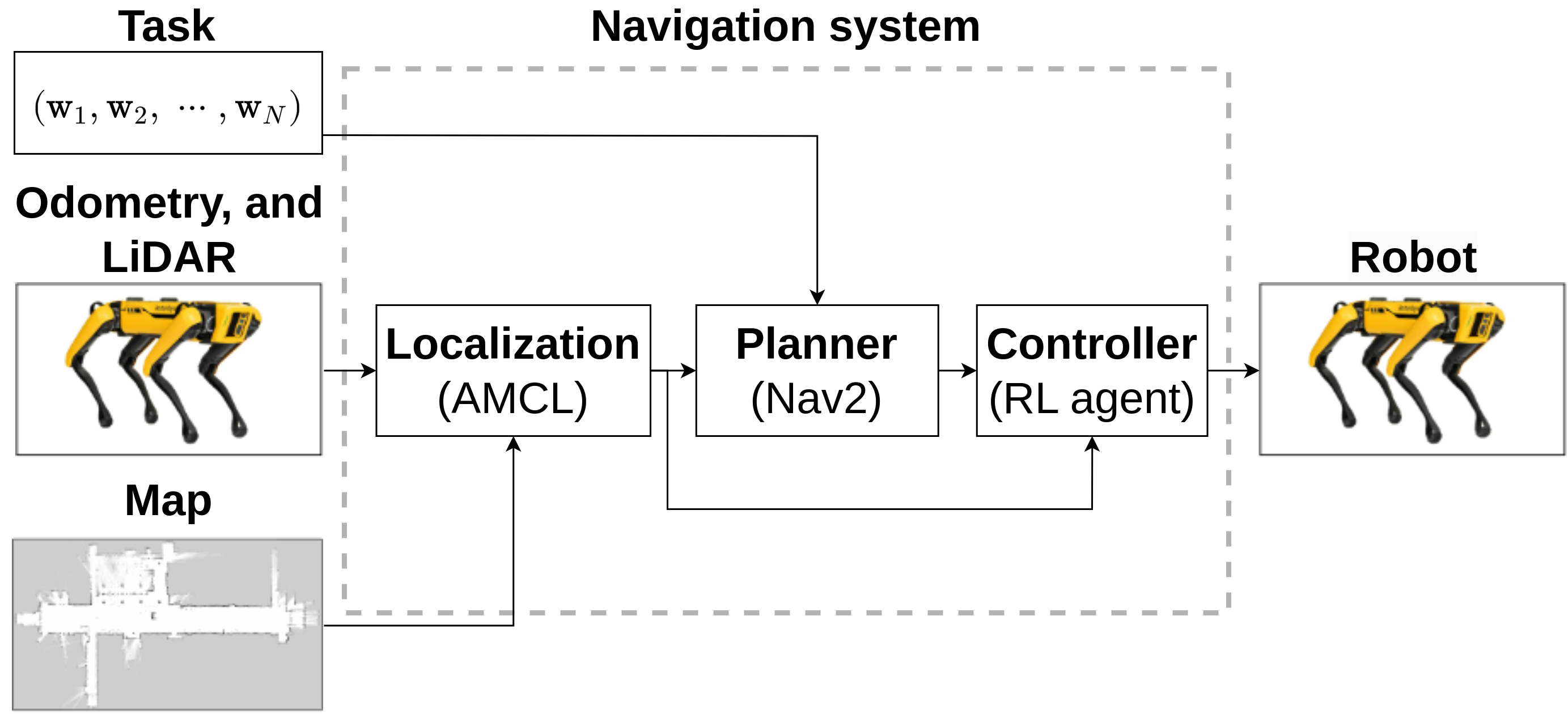}
    \caption{The navigation system used for the surveillance task.}
    \label{fig:navigation_system}
\end{figure}

The navigation system, illustrated in Fig.~\ref{fig:navigation_system} is comprised of the following sub-modules:

\begin{figure}
    \begin{subfigure}{1.0\linewidth}
    \centering
    \includegraphics[width=0.85\linewidth, trim=0.1cm 0.2cm 0.2cm 0.1cm, clip]{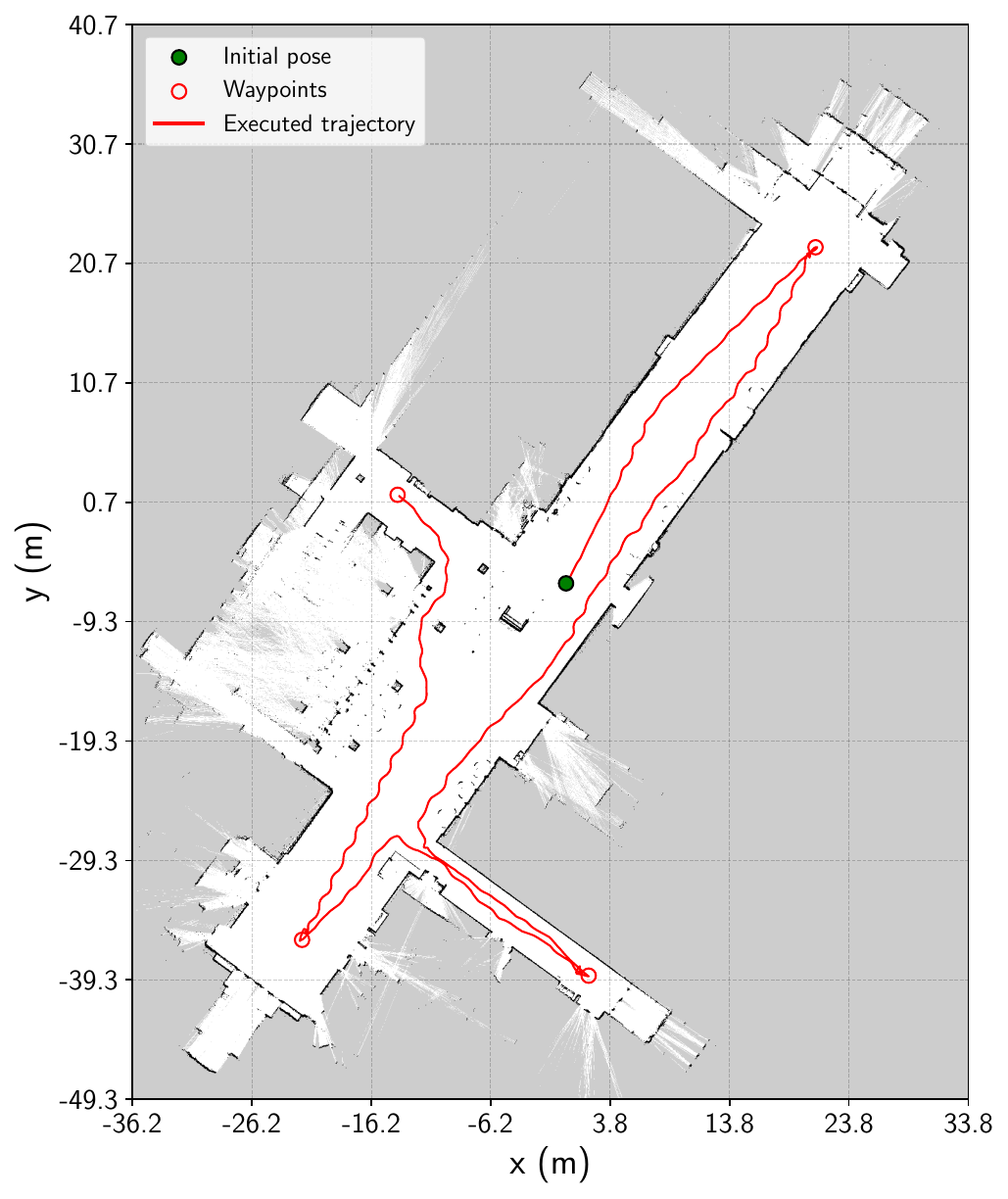}
    \caption{}
    \label{fig:executed_path_a}
    \vspace{1em} 
    \end{subfigure}
    \begin{subfigure}{1.0\linewidth}
    \centering
    \includegraphics[width=1.0\linewidth]{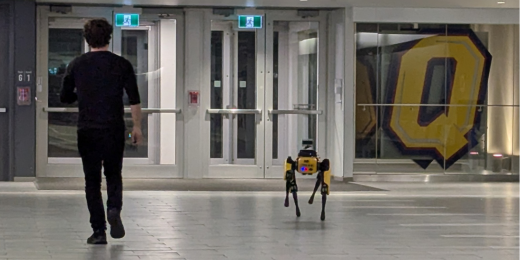}
    \caption{}
    \label{fig:executed_path_b}
    \end{subfigure}
    \caption{(a) Executed trajectory (in red) using the trained RL agent on the test site map represented as an occupancy grid (white is free space, black is occupied, and grey is unknown) and (b) photo of the robot near the first waypoint.}
    \label{fig:executed_path}
\end{figure}

\subsubsection{Inputs}
The inputs of the navigation system include the task, represented as a sequence of waypoints $(\mathbf{w}_1,\mathbf{w}_2, \cdots, \mathbf{w}_N)$, the map of the building, obtained using SLAM with a resolution of $0.05$ m, and the robot sensors comprised of the internal odometry and distance measurements using a VLP-16 Velodyne LiDAR. 

\subsubsection{Localization}
The localization system consisted of the Adaptive Monte Carlo Localization (AMCL) library available for ROS that implements an adaptive particle filter to estimate the robot's pose.

\subsubsection{Planner}
The state lattice planner available in the Nav2 ROS library was used to generate the paths. This algorithm uses motion primitives based on the robot's kinematics to find a collision-free path between two configurations. This planner is capable of generating motion primitives for both car-like and omnidirectional robots. However, the omnidirectional version produced paths that required the robot to move sideways, which is the slowest direction, for the majority of the task execution. In contrast, the car-like version generated more suitable paths, wherein the robot faced forward for most of the task. A small steering radius of $0.1$ m was chosen to allow the robot to turn in place. The state lattice planner outputs paths with the same resolution as the map, $0.05$ m. To account for the RL agent tolerance of $0.3$ m, the path was under-sampled to contain $1$ m segments.

\subsubsection{Controller}
The RL agent controller, trained through the pipeline, followed the planned path by sequentially controlling the robot toward each point in the path. This was achieved by representing the path as a sequence of sub-goals. Once the robot reaches a sub-goal within the specified tolerances $\epsilon_p$ and $\epsilon_\theta$, the robot is sent to the next sub-goal in the path.

\subsection{Application to Surveillance}

Fig.~\ref{fig:executed_path} shows Spot's path during the surveillance task. The robot was positioned near the centre of the space, indicated by the green circle, with waypoints selected near the extremities of each side of the building, indicated as red circles.

During the execution, the robot's speeds ($a_x$, $a_y$, $a_{\theta}$) were limited to $1$~m$/$s. This limit was added as a security measure to keep the robot within the speed limits of the identified model. The total length of the path executed was 197~m, and the robot completed the task in $201$~s. This results in an average speed of $0.98$~m$/$s. A slightly lower average speed is expected since the robot has to change directions at each waypoint.

While the robot completed the task with the expected tolerances and speeds, some oscillations were observed in the path. These oscillations are particularly noticeable near the first and third waypoints. Although the exact causes are still unclear, they were not present in Fig.~\ref{fig:policy_results_c} experiment. Therefore, these oscillations probably appeared due to the integration of the RL agent and the navigation system. One possible reason for some of the oscillations is a sudden change in orientation in the planned path. Additionally, the RL agent may be overcompensating due to larger variations in the estimated pose than what it was trained to handle. Further experiments are needed to understand better these oscillations and how to mitigate them.
\section{Conclusion}
\label{sec:conclusion}

This paper introduces a Reinforcement Learning (RL) pipeline that incorporates multiple stages of training, from simulation to real-world robotic applications. Our pipeline includes two levels of simulation complexity for sequential training of the RL agent, along with iterative feedback paths that support an iterative refinement of the policy. By incorporating our pipeline and applying techniques such as system identification, domain randomization, and curriculum learning, it is possible to reduce the reality gap and facilitate safe and efficient policy training in robotics. We also present a case study using the Boston Dynamics Spot robot in a surveillance application, demonstrating the practical value of our pipeline through a successful deployment of an RL agent.

\balance
\bibliographystyle{ieeetr} 

\end{document}